\documentclass[letterpaper]{article} 
\usepackage{aaai23}  
\usepackage{times}  

\usepackage{helvet}  
\usepackage{courier}  
\usepackage[hyphens]{url}  
\usepackage{graphicx} 
\usepackage{natbib}  
\usepackage{caption} 
\usepackage{algorithm}
\usepackage{algorithmic}

\usepackage{newfloat}
\usepackage{listings}
\DeclareCaptionStyle{ruled}{labelfont=normalfont,labelsep=colon,strut=off} 
\floatstyle{ruled}
\newfloat{listing}{tb}{lst}{}
\floatname{listing}{Listing}
\title{RFC-Net: Learning High Resolution Global Features for Medical Image Segmentation on a Computational Budget (Student Abstract)}
\author {
    Sourajit Saha\textsuperscript{\rm 1},
    Shaswati Saha\textsuperscript{\rm 2},
    Md Osman Gani\textsuperscript{\rm 3},
    Tim Oates\textsuperscript{\rm 4},
    David Chapman\textsuperscript{\rm 5}
}
\affiliations {
    \textsuperscript{\rm 1,\rm 4} Department of Computer Science and Electrical Engineering, University of Maryland Baltimore County, MD 21250, USA\\
    \textsuperscript{\rm 2, \rm 3} Department of Information Systems, University of Maryland Baltimore County, MD 21250, USA\\
    \textsuperscript{\rm 5} Department of Computer Science, University of Miami, FL 33124, USA\\
    \{\textsuperscript{\rm 1}ssaha2, \textsuperscript{\rm 2}ssaha3, \textsuperscript{\rm 3}mogani, \textsuperscript{\rm 4}oates\}@umbc.edu, \textsuperscript{\rm 5}dchapman@cs.miami.edu
}

\usepackage{bibentry}

\begin{document}

\maketitle

\begin{abstract}
Learning High-Resolution representations is essential for semantic segmentation. Convolutional neural network (CNN) architectures with downstream and upstream propagation flow are popular for segmentation in medical diagnosis. However, due to performing spatial downsampling and upsampling in multiple stages, information loss is inexorable. On the contrary, connecting layers densely on high spatial resolution is computationally expensive. In this work, we devise a Loose Dense Connection Strategy to connect neurons in subsequent layers with reduced parameters. On top of that, using a m-way Tree structure for feature propagation we propose Receptive Field Chain Network (RFC-Net) that learns high-resolution global features on a compressed computational space. Our experiments demonstrates that RFC-Net achieves \textit{state-of-the-art} performance on \textit{Kvasir} and \textit{CVC-ClinicDB} benchmarks for Polyp segmentation. Our code is publicly available at \textbf{\textit{github.com/sourajitcs/RFC-NetAAAI23}}.
\end{abstract}

\section{Introduction}
Decreasing spatial resolution in CNN's forward propagation engenders difficulty in learning high-resolution global features which affects pixel-wise image segmentation quality. Preserving spatial resolution therefore, has been a consistent design choice among a number of high-precision CNN models \cite{wang2020deep} proposed in recent times. However, such design choices \cite{han2022seethroughnet} lead to computational overhead due to having multiple ResNet and transformer blocks running in parallel at high-resolution (spatial). Despite their high precision, training such models are often challenging in the presence of constraints in computational budget and n-dimensional imagery where $n \in \{\mathcal{Z}, n>2\}$. To mitigate this bottleneck, we propose \textbf{R}eceptive \textbf{F}ield \textbf{C}hain \textbf{Net}work \textbf{(RFC-Net)} by means of devising the following:
\begin{equation}
  \|\Theta_{SDCS}\|_{l}^{l+1} = d_{l+1} (\mathbf{[k^{2}]} d_{l} + \mathbf{d_{l+1}})
  \label{eqn:param_SDCS}
\end{equation}
\begin{itemize}
    \item Introduce \textbf{L}oose \textbf{D}ense \textbf{C}onnection \textbf{S}trategy \textbf{(LDCS)} for reducing the number of parameters.
    \item Design a \textbf{m-way Tree structure} to learn features through a chain constituting all possible combination of different receptive fields from a selected set of kernels.
\end{itemize}
\begin{figure}[t]
\centering
\includegraphics[height=3.95cm, width=\columnwidth]{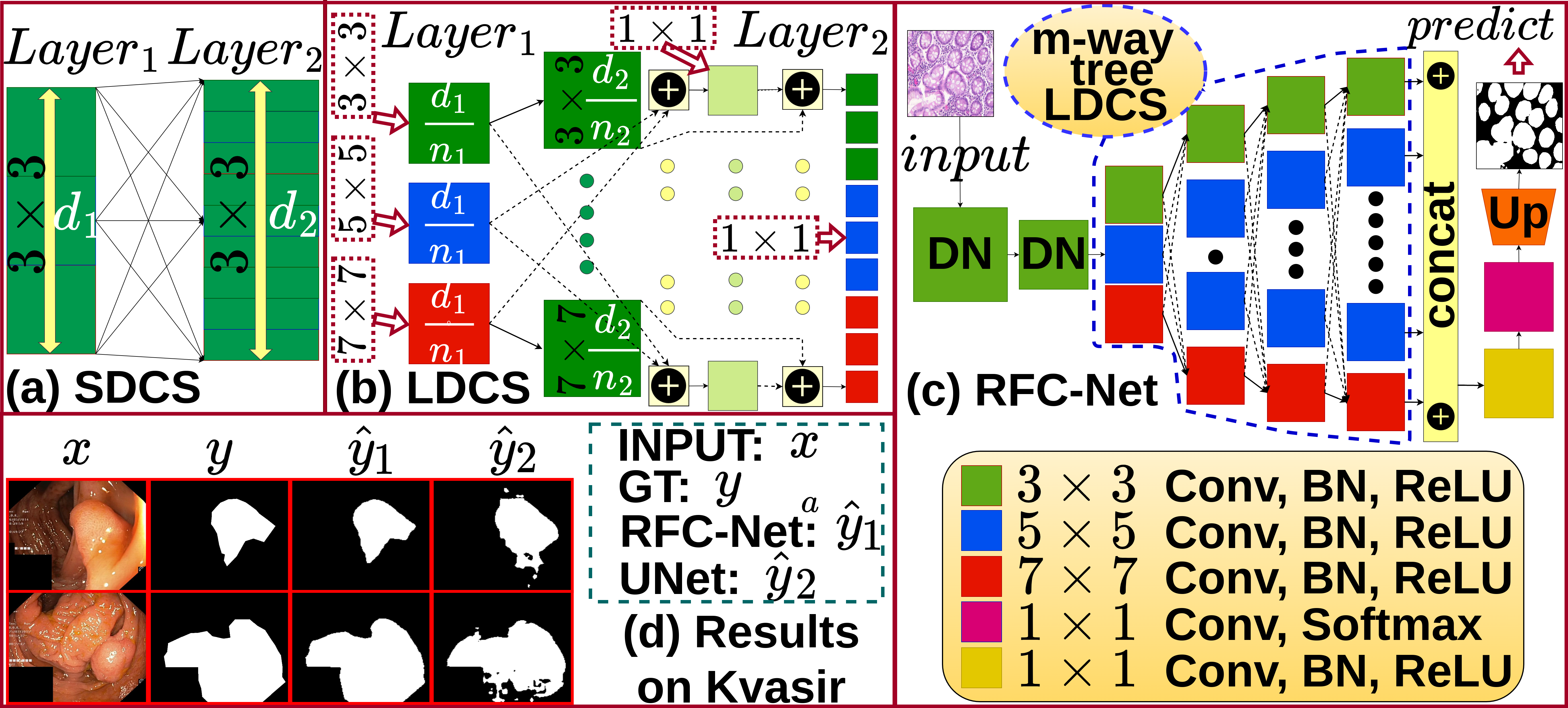} 
\caption{Enhanced view recommended. \textbf{(a):} Conventional Strong Dense Connection Strategy (SDCS). \textbf{(b):} Proposed Loose Dense Connection Strategy (LDCS). \textbf{(c):} Proposed RFC-Net Architecture: Expanding with m-way tree on uncompressed resolution with LDCS (DN: Downsample). \textbf{(d):} Comparison between the quality of RFC-Net and UNet's prediction $(\hat{y})$ mask on Kvasir dataset (GT: Ground Truth).}
\label{fig:illustration}
\end{figure}
\section{Methodology}
As constituent of our proposed RFC-Net we first introduce Loose Dense Connection Strategy (LDCS) which in comparison with conventional Strong Dense Connection Strategy (SDCS) aids in reducing the number of parameters as illustrated in Figure ~\ref{fig:illustration}(a-b). We define strong connection between two neurons as being filtered through a $k \times k$ convolutional kernel $(\forall k>1)$ since such connections enable learning spatial correlations whereas, we define loose connection as using a $k \times k$ kernel $(\forall k=1)$ and this operation is equivalent to that of a linear layer with no propagation of spatial information about neighbouring pixels. Using $k \times k$ kernels, conventional SDCS \cite{ronneberger2015u} strongly $(k>1)$ connects every neurons from $l^{th}$ layer to accrue each neuron in the $(l+1)^{th}$ layer and Equation ~\ref{eqn:param_SDCS} denotes the number of parameters $(\|\Theta\|)$ to constitute the entire $(l+1)^{th}$ layer where $d_{l}$ is the total number of neurons in layer $l$, $l=\{1,2,..,L\}$ and L is the total number of layers. However, in LDCS we split the number of neurons in $l^{th}$ layer to $n_{l}$ groups. Using our proposed LDCS, we first of all strongly $(k>1)$ connect one of those $n_{l}$ groups in $l^{th}$ layer to one of those $n_{l+1}$ groups in $(l+1)^{th}$ layer. As illustrated in Figure ~\ref{fig:illustration}(b), we then concatenate the other ($d_{l}-n_{l}$) groups in $l^{th}$ layer and loosely $(k=1)$ connect them to the previously chosen $n_{l+1}^{th}$ group in $(l+1)^{th}$ layer where we performed the strong connection with $n_{l}^{th}$ group. Finally, we concatenate the strongly and loosely connected neurons to the aforementioned $n_{l+1}^{th}$ group and pass through one more $(1\times1)$ kernel to construct the final $n_{l+1}^{th}$ group in $(l+1)^{th}$ layer. Equation ~\ref{eqn:param_LDCS} depicts the number of parameters $(\|\Theta\|)$ required to construct the entire $(l+1)^{th}$ layer using LDCS. The computation reduction factors driven by LDCS are \textbf{highlighted} in Equation ~\ref{eqn:param_SDCS} and ~\ref{eqn:param_LDCS}. 
\begin{equation}
  \|\Theta_{LDCS}\|_{l}^{l+1} = d_{l+1} (\left[\mathbf{\frac{k^{2}}{n_{l}}}+\frac{n_{l}-1}{n_{l}}\right] d_{l} + \mathbf{\frac{d_{l+1}}{n_{l+1}}})
  \label{eqn:param_LDCS}
\end{equation}
\begin{equation}
  x_{i}^{l+1} = \mathcal{F}_{1\times1}\mathbf{(}\mathcal{F}_{k\times k}(x_{\lceil i/m \rceil}^{l}) \oplus \mathcal{F}_{1\times1}(\oplus_{j \in n_{l}}x_{j\neq \lceil i/m \rceil}^{l})\mathbf{)}
  \label{eqn:propagation_forward}
\end{equation}

Secondly, we introduce a m-way Tree structure into the network and the number of groups at layer $l$, $n_{l}=m^{l}$. We repeatedly deploy m kernels $(k>1)$ of different receptive fields throughout each layers across the network as shown in Figure ~\ref{fig:illustration}(c). Equation ~\ref{eqn:propagation_forward} shows how connecting these groups across layers with LDCS using a m-way Tree structure thus facilitates the propagation of features through a chain constituting all possible combination of different receptive fields with different values of $k$, where $\mathcal{F}_{k\times k}$ is a convolution kernel of size $k\times k$ and $x_{i}^{l}$ denotes $i^{th}$ intermediate feature representation at layer $l$. However, connecting only one group from a layer to that of the next one (considering only the $\mathcal{F}_{k\times k}$ portion in Equation ~\ref{eqn:propagation_forward}) leads to amassing $m^{L}$ number of week-performing segregated networks due to lack of information exchange. On the contrary, connecting all of the groups using the conventional SDCS strategy will result in a combinatorial explosion in computation. We observe that, replacing $n_{l}$ in Equation ~\ref{eqn:param_LDCS} with $m^{l}$ exponentially reduces the computation constraints. Therefore, juxtaposing m-way Tree structure with different receptive fields and LDCS enables Receptive Field Chain Network termed as RFC-Net (Figure ~\ref{fig:illustration}(c)) to learn high-resolution aware features within computational budget. Finally, learning features on a hierarchical space of cascaded receptive fields without severe spatial downsampling (we only downsample twice before passing through the m-way tree, as illustrated in Figure ~\ref{fig:illustration}(c)) helps RFC-Net learn more robust global features.
\section{Experiments, Results And Observations}
We evaluate RFC-Net on three binary segmentation benchmarks- (1) Kvasir SEG Polyp Dataset (Gastrointestinal Disease Detection, train:test=850:150), (2) GlaS Dataset (Gland Segmentation in Colon Histology, train:test=132:33), (3) CVC-ClinicDB (Polyp segmentation from colonoscopy video frames, train:test=521:91). We use SGD optimizer with momentum=0.9, weight decay=0.0005. Furthermore, we train (without data augmentation) all our models on the benchmarks with Online Hard Example Mining Cross Entropy (OHEM CE) loss (threshold=0.7). Firstly, we train Kvasir SEG Polyp Dataset (resized to 200$\times$200, batch size of 4) for 160 epochs (base learning rate=0.01, step size=45, converges in 77 epochs). Secondly, we train GlaS Dataset (resized to 300$\times$300, batch size of 2) for 500 epochs (base learning rate=0.01, step size=60, converges in 259 epochs). Thirdly, we train CVC-ClinicDB Dataset (resized to 200$\times$300, batch size of 3) for 160 epochs (base learning rate=0.01, step size=45, converges in 85 epochs). We ran all of our experiments on PyTorch using one Nvidia RTX 3090 and one Nvidia RTX 3070 GPU.

In Table ~\ref{tab:results_and_ablation}, we narrate how RFC-Net performs in comparison to existing heavier and lighter models. We observe that while reducing parameters RFC-Net$^{a}$ outperforms the existing models in Kvasir and CVC-ClinicDB benchmarks and exhibits comparable performance on GlaS dataset. We further perform an ablation study (Table ~\ref{tab:results_and_ablation}), to inspect the efficacy of our model. We notice that RFC-Net$^{b}$ with even lesser parameters outperforms the existing models on kvasir and CVC-Clinic DB benchmark. We further observe that, it is the exploitation of the m-way tree structure by RFC-Net$^{a}$ for using enhanced receptive field that leads to a sharp performance gain (RFC-Net$^{b}\mapsto$RFC-Net$^{a}$, RFC-Net$^{d}\mapsto$RFC-Net$^{c}$). Additionally, the performance of RFC-Net's lightweight versions (RFC-Net$^{c}$, RFC-Net$^{d}$) is substantially higher than that of ESPNet-C which is another existing lightweight CNN for segmentation. Figure ~\ref{fig:illustration}(d) further depicts the higher quality (in comparison to UNet) of prediction masks produced by RFC-Net. Being computationally inexpensive, RFC-Net can be further applied to 3D and 4D image segmentation for \textit{Computer Aided Diagnosis}.
\begin{table}[t]
\centering
\resizebox{.99\columnwidth}{!}{
\begin{tabular}{l|cc|ccc}
    \hline
     & \multicolumn{2}{c|}{\textit{Computation}} & \multicolumn{3}{c}{\textit{Performance (mIOU \%)}} \\ \hline
    \textit{Model} & \textit{Params} & \textit{GFLOPs} & \textit{Kvasir} & \textit{GlaS} & \textit{CVC-DB} \\ \hline
    U-Net & 07.76M & 30.75B & 74.39 & 83.51 & 75.91 \\ \hline
    U-Net++ & 09.04M & 34.91B & 75.95 & 84.02 & 79.27 \\ \hline
    ResUNet & 13.04M & 43.56B & 77.77 & \textbf{85.67} & 81.33 \\ \hline 
    ESPNet-C & 00.41M & 02.49B & 75.95 & 65.43 & 72.53 \\ \hline
    \textbf{RFC-Net$^{a}$} & \textbf{05.76M} & \textbf{18.13B} & \textbf{81.31} & 77.88 & \textbf{85.90} \\ \hline
    RFC-Net$^{b}$ & 04.49M & 14.03B & 79.15 & 75.34 & 83.34 \\ \hline
    RFC-Net$^{c}$ & 00.39M & 01.27B & 76.41 & 75.49 & 79.51 \\ \hline
    RFC-Net$^{d}$ & 00.28M & 00.91B & 73.24 & 66.17 & 77.68 \\ \hline
\end{tabular}
}
\caption{Results on Kvasir, GlaS, CVC-ClinicDB Datasets. GFLOPs are calculated on $224\times 224$ RGB image. RFC-Net$^{a}$:[m=3,k=3,5,7], RFC-Net$^{b}$:[m=3,k=3,3,3], RFC-Net$^{c}$:[m=2,k=3,5], RFC-Net$^{d}$:[m=2,k=3,3].}
\label{tab:results_and_ablation}
\end{table}
\bibliography{aaai23}
\end{document}